\documentclass[a4paper,twoside]{article}

\usepackage{epsfig}
\usepackage{fancyheadings}
\usepackage{array}
\usepackage{multirow}
\usepackage{calc}
\usepackage{amssymb}
\usepackage{amstext}
\usepackage{amsmath}
\usepackage{amsthm}
\usepackage{multicol}
\usepackage{pslatex}
\usepackage{apalike}
\usepackage{xcolor}
\usepackage{todonotes}
\usepackage{amsmath,array,graphicx}
\usepackage{kantlipsum}
\usepackage{multirow}
\usepackage{enumitem}
\usepackage{algorithm}
\usepackage{algorithmic}
\usepackage{svg}
\usepackage{graphicx}
\usepackage{subfig}
\usepackage{glossaries}
\usepackage{hyperref}
\usepackage[compact]{titlesec}
\newglossaryentry{hci}{name=HCI,description=Human Computer Interaction}
\newglossaryentry{dnns}{name=DNNs,description=Deep Neural Networks}
\newglossaryentry{dnn}{name=DNN,description=Deep Neural Network}
\newglossaryentry{wslf-lw}{name=WSLF-LW,description=Weighted Score-Level Fusion with Learning Weights}
\newglossaryentry{sh}{name=SH,description=Smart Homes}
\newglossaryentry{gpd}{name=GPD,description=Geometric Pose Descriptor}
\newglossaryentry{cnns}{name=CNNs,description=Convolutional Neural Networks}
\newglossaryentry{vfoa}{name=VFOA,description=Visual Focus of Attention}
\newglossaryentry{svm}{name=SVM,description=Support Vector Machines}
\newglossaryentry{logit}{name=Logit,description=Logistic regression}
\newglossaryentry{mlp}{name=MLP,description=Multilayer Perceptron}
\usepackage{SCITEPRESS}     % Please add other packages that you may need BEFORE the SCITEPRESS.sty package.
\newcommand\MyBox[2]{
  \fbox{\lower0.75cm
    \vbox to 1.7cm{\vfil
      \hbox to 1.7cm{\hfil\parbox{1.4cm}{#1\\#2}\hfil}
      \vfil}%
  }%
}
\titlespacing*{\section}{0pt}{*0}{*0}

\begin{document}

%\setkeys{Gin}{draft}
%\title{Measuring the Attention Level of a User Through Contextual Features in the Home}
%\title{Vision-Based Estimation of Attention Level in the Home\\Using Subjective Annotations}
\title{Subjective Annotations for Vision-Based Attention Level Estimation}

\thispagestyle{myheadings}
\markright{\href{http://www.visapp.visigrapp.org/}{VISAPP 2019}\hspace{3.5cm} February 25-27, 2019, Prague, CZ\hfill}
\pagestyle{myheadings}
\markright{\href{http://www.visapp.visigrapp.org/}{VISAPP 2019}\hspace{3.5cm} February 25-27, 2019, Prague, CZ\hfill}
\author{\authorname{Andrea Coifman\sup{1}, P\'{e}ter Rohoska\sup{1,3}, Miklas S. Kristoffersen\sup{1,2},\\Sven E. Shepstone\sup{2}, and Zheng-Hua Tan\sup{1}}
\affiliation{\sup{1}Department of Electronic Systems, Aalborg University, Denmark}
\affiliation{\sup{2}Bang \& Olufsen A/S, Struer, Denmark}
\affiliation{\sup{3}Continental Automotive, Budapest, Hungary\thanks{The presented work was done in affiliation with \sup{1} but the affiliation of the author changed to \sup{3} during the submission. }.}
%\email{\{msk, zt\}@es.aau.dk}
}

\keywords{Attention Level Estimation, Natural HCI, Human Behavior Analysis, Subjective Annotations}

\abstract{Attention level estimation systems have a high potential in many use cases, such as human-robot interaction, driver modeling and smart home systems, since being able to measure a person's attention level opens the possibility to natural interaction between humans and computers. The topic of estimating a human's visual focus of attention has been actively addressed recently in the field of HCI. However, most of these previous works do not consider attention as a subjective, cognitive attentive state. New research within the field also faces the problem of the lack of annotated datasets regarding attention level in a certain context. The novelty of our work is two-fold: First, we introduce a new annotation framework that tackles the subjective nature of attention level and use it to annotate more than 100,000 images with three attention levels and second, we introduce a novel method to estimate attention levels, relying purely on extracted geometric features from RGB and depth images, and evaluate it with a deep learning fusion framework. The system achieves an overall accuracy of 80.02\%. Our framework and attention level annotations are made publicly available.}

\onecolumn \maketitle \normalsize \vfill
% \todo{make registration}
% \todo{contact PANDORA ppl}
% \todo{put paper on arxiv - check with VISAPP requirements}
% \todo{justify RGB}
\section{\uppercase{Introduction}}
\label{sec:introduction}
% \noindent Your paper will be part of the conference proceedings
% therefore we ask that authors follow the guidelines explained in
% this example in order to achieve the highest quality possible
% \cite{Smith98}.

% Be advised that papers in a technically unsuitable form will be
% returned for retyping. After returned the manuscript must be
% appropriately modified.

Natural interaction has become an essential element in Human-Computer Interaction (\gls{hci}) over the past years. In the context of smart homes, the integration of intelligent systems into the home environment has improved the natural and effective communication between humans and home appliances. However, providing a natural interaction between smart devices and humans without the use of a physical device, e.g. a remote control, is still a challenging task. Vision-based systems are able to provide a more efficient and natural way of communication. As the years pass, these systems are starting to be more frequently applied into our daily living and homes.
Although interaction between humans and computers is already well established in our homes, ways to make these systems more intelligent still draw much research interest. One common improvements is the integration of an intelligent system to give recommendations to the users \cite{gomez-uribe_netflix_2016,nararajasivan_location_2016} based on learned interests. Most studies are based on integrating context-aware devices to track human behavior and routines \cite{tamdee_context-aware_2018}. In this paper, we propose to use attention levels as an effective means to improve natural \gls{hci} in the home by measuring and classifying the attention level of a user. Attention can be defined as the focus of cognitive resources on a task, ignoring sources of distraction \cite{zaletelj_predicting_2017}. In other words, \textit{attention is related to whether a user is focused or not on the task at hand, e.g. the interaction with a device}.

Quantitative techniques for measuring attention levels rely on the use of different sensors depending on the type of measurement: direct or indirect \cite{mancas_how_2016}. Direct measurements include brain responses e.g. by using EEG \cite{murthy_g.n._cognitive_2014} techniques, among others. However, extracting data from these sensors is expensive, intrusive, and uncomfortable for a natural communication. To address this problem, indirect measurement of attention levels has been studied through non-intrusive observation of users' behaviors, e.g. by using eye-trackers or head pose estimators \cite{asteriadis_importance_2011,masse_tracking_2017}. Indirect measurements do not require the use of a physical sensor attached to the user, which makes the interaction with the system more natural. Most works on estimating attention levels through indirect measurements rely on the gaze direction of a user. This is due to the fact that the eye gaze is highly correlated with what a user is interested in \cite{zaletelj_predicting_2017}. Existing datasets that provide annotations rely on objective frameworks e.g. calculating the direction of gaze or movements of eyes among other techniques \cite{kar_review_2017}. % In the field of natural \gls{hci}, most of the existing datasets that provide annotations rely on a non-subjective framework for the task of attention estimation.% \todo[inline]{This claim needs further attention, and subjective vs non-subjective (normally you would call that objective) needs to be explained, so the reader knows what we are talking about. We need to be careful here, I would assume that objective annotations are much more reliable than subjective. So why do we really use subjective annotations?? The non-subjective frame-works you mention, are those head pose / eye gaze systems, where the target is simply the degrees/angles? Then we should make that clear and tell why we think it's a better idea to use human intuition/subjective annotations.} 

This paper contributes to the field of \gls{hci} by providing subjective attention level annotations to an existing publicly available dataset, Pandora \cite{borghi_poseidon:_2017}, that contains visual recordings of users. A subjective label of attention level refers to the fact that each of the labelers bases their labeling decision on how they perceive each user's attention level in the recordings, by taking the established definition of attention level into account. Based on the annotations, we further present a novel method that is able to measure attention level automatically. Our framework and attention level annotations are made publicly available\footnote{http://kom.aau.dk/$\sim$zt/online/SubjectiveAnnotations/}. The remainder of this paper is organized as follows: Section \ref{s:related} reviews previous state-of-the-art methods for estimating attention alongside existing datasets; Section \ref{s:annotations} explains the process of annotating the dataset; Section \ref{s:baseline} presents a baseline for estimating attention with subjective attention annotations; in Section \ref{sec:experiments} we present our experiments and quantitative results for the previously built baseline.

\section{\uppercase{Related Works and Motivations}}\label{s:related}

% \todo[inline]{I would make geometric features a subsection of vision-based attention estimation methods. Maybe we need something about end-to-end (raw images-CNN/DNN-attention), and why we use geometric features. Again, be careful not to say that geometric features are indisputably better (many would think otherwise), so it's probably more about explainability (or what is your motivation??).}
% \todo[inline]{I'll revisit this section, when citations have been added and the dataset section has been formed.}

\subsection{Vison-based Attention Estimation}\label{ss:methods}

First works on addressing attention estimation problems were based on predicting head pose or direction of eye gaze, which are considered to be high level descriptors of attention. This is due to the fact that, normally, the position of the head and the gaze direction are highly correlated with the subject of the user's interest or, more applied to this paper, what caught the user's attention \cite{jariwala_robust_2016,zaletelj_predicting_2017}. Majority of studies focused on determining the Visual Focus of Attention (\gls{vfoa}) for users' attention estimation. The \gls{vfoa} denotes the target of what a user is looking at and it is mostly determined by the combination of a user's eye gaze and head pose dynamics \cite{masse_tracking_2017}. The \gls{vfoa} characterizes a perceiver and target pair. It may be defined either by the line from the perceiver’s face to the perceived target, or by the perceiver’s direction of sight or gaze direction. However, estimating the attention level of a user based only on eye gaze can be problematic in the home context, due to the lack of detection range and the sensor's low resolution.

In this work, we formulate the attention level estimation as a classification problem to be solved in a supervised learning framework, since we have subjective annotations of attention levels on a dataset. To this end, we propose a set of geometric features as an effective representation of high-level features that will serve as descriptors of the attention level of a user. The set of representative features consist of face, head, and body points, distances, and angles that describe the eye gaze direction together with head and body orientation of the user. There are a number of advantages of using the proposed features in a supervised framework. First, it includes more than eye gaze and pose information and thus is more descriptive. Second, it avoids deploying separate eye gaze and pose estimation systems and thus potentially reduces the complexity of the system. Finally, the system works even when eyes are not visible as multiple features are involved in the estimation of attention levels.

% There have been several studies on trying to estimate attention based on visual attributes of a user. Visual Focus of Attention (\gls{vfoa}) specializes in estimating user's attention based on visual cues by calculating user's gaze direction . However, estimating attention based only on the user's gaze can be problematic in some contexts. In the home context (e.g. in the living room), the eyes of the user are sometimes not visible e.g. when the intelligent system, such as TV, is located far from the user. To address this problem, further works on improving \gls{vfoa} integrate head pose information to the system, independent to the gaze direction {\color{red}citation}.

% \todo[inline,color=red]{introduction to geometric features}
Many prior works in the field of \gls{hci} aim to use lower-level geometric features (e.g. locations, distances or angles between locations) to predict and classify higher-level features (e.g. direction of eye-gaze, body pose or head pose) for attention level estimation. Chen \textit{et al.} \cite{chen_novel_2016} evaluated five dimensional feature vectors, containing geometric features of detected joints to evaluate joint motions for action recognition. Yun \textit{et al.} \cite{yun_two-person_2012} evaluates two-person interaction based on a wide variety of geometric body features, such as joint keypoints and distances, and joint-to-plane distances. Mass\'{e} \textit{et al.} \cite{masse_tracking_2017} propose a framework where correlation between head pose and eye gaze is used to estimate the \gls{vfoa}.
The authors of some of these works also address the importance of these features to estimate attention in the field of \gls{hci}. To this extent, we decide to use geometric features and the corresponding relation between these features to evaluate the level of attention of a user.

\subsection{Datasets and Annotation Methods}\label{ss:previous}

Different approaches have been proposed for generating attention estimation datasets and corresponding annotations. The most common way for researchers is to record their own datasets and hand label the images with information regarding the direction of the gaze \cite{tseng_camera-based_2017} or shifts of attention between targets over the images in the dataset \cite{steil_forecasting_2018}. All of the previously mentioned datasets and their annotations can be used to estimate the focus of attention, through the eye gaze and head pose labels. However, there are no existing information on the level of attention, even though, most of the datasets can be re-labeled with attention level annotations.
Asteriadis \textit{et al.} \cite{asteriadis_importance_2011} proposed a dataset similar to ours. In their work, the authors hand-labeled the Boston University \cite{boston} dataset using similar approaches regarding attention levels. First, the images were labeled regarding annotator's perception of attention over the subject in the image, with two levels - '0' and '1' regarding attentive or non-attentive state. Second, an average decision rule was applied in order to evaluate the agreement between annotators. In contrast to the authors' work, we hand-label the images with three levels of attention - '0', '1', and '2' regarding low, medium, and high attentive state. For evaluating the level of agreement between annotators, we apply a "majority" decision rule which selects the final label as the one that has more than half of the votes from the labelers. In this way, we ensure a fair decision. Furthermore, we evaluate geometric features to address the problem of estimating attention level of a user.

% \todo[inline]{Would be cool with an ending similar to the previous sections, i.e. a few lines about what makes our work stand out.}

\section{\uppercase{Subjective Annotations}}\label{s:annotations}

In this section, the publicly available dataset and the approach towards hand-labeling the 130,889 RGB images with subjective attention annotations are explained. A decision rule is later applied in order to evaluate the agreement between the labels from different labelers for each video frame.

\subsection{Original Data and Reason of Use}

Borghi \textit{et al.} \cite{borghi_poseidon:_2017} first introduced the Pandora dataset in 2016 as a part of research project in the automotive field for head center localization, head pose, and shoulder pose estimation. The dataset contains 110 sequences of 22 different subjects performing constrained and unconstrained driving-like movements. The data was collected using a Microsoft Kinect One device, which acquires the upper-body part of each of the subjects. The authors stated that, in this way, the position of the sensor would simulate as if it was located in the car's dashboard. In this paper we follow this idea considering the lack of attention level datasets in the context of the living room. The position of the sensor and the captured upper-body simulates very well the idea of capturing attention level of a user from a smart system, such as smart TVs.

\subsection{Annotation Method}\label{ss:annotations}

In the present paper, a version\footnote{The authors are still updating the Pandora dataset with new RGB and depth images. Last noticed version updated on 29th of May, 2018.} of the dataset that contains 261,778 RGB (1920x1080 pixels) and depth (512x424) images was used. The dataset contains five different sequences of constrained (three) and unconstrained (two) actions from 20 different subjects. Contrary to previous datasets explained in Section \ref{ss:previous} with annotations based on shifts of attention, we propose a frame-by-frame approach, in order to better understand the context and different situations. To this end, the data was re-structured as unique sequences for each of the 20 subjects, with jointly constrained and unconstrained movements, ending up with an average of 6,544 RGB frame images per subject. The frames were manually annotated with subjective annotations of attention level by five annotators: four were used for labeling and one for checking. A subjective label of attention level refers to the fact that each of the labelers based their labeling decision on their personal feelings and opinion. An objective definition of attention (see Section \ref{sec:introduction}) was provided to the labelers beforehand, along with the definition of each attention level: 

\begin{enumerate}
\item['0'] Low attention level: The subject was not paying attention to the task at hand. 
\item['1'] Mid attention level: The subject was partially paying attention to the task at hand.
\item['2'] High attention level: The subject was paying full attention to the task at hand.
\end{enumerate}

% An image-labeling software tool was built, which takes as input a set of images from the dataset, and frame-by-frame the annotators could select among three levels of attention. 
% \todo[inline]{Some reviewers will definitely ask: Who are the labelers, how did you recruit them, how did you instruct them, what was the objective definition? Try to do it short and concise, just enough to satisfy the curiosity of the reviewers.}
% \todo[inline]{Where is it mentioned that there are four annotators? Also see the commented todo above this todo.}

\subsection{Decision Rule and Agreement}

The total number of annotations was 523.556 (four labels per frame). In order to measure the agreement between the annotators, a decision rule was determined. In this way, every frame would have a unique label. The decision rule aim to assign the most fair label to each of the frames. A first decision rule was applied over the annotations by selecting the alternative that had more than half of the votes (majority rule). In this way, votes were treated equally. Having four annotations per frame, majority was settled when three or more labels were equal, ending up with an agreement over all the annotations from the dataset of 79.7\%. In order to increase the percentage of agreement between the annotators, a checker labeler was introduced to annotate the unsolved frames. The annotations procedure for the checker was the same as for the rest of the annotators (explained in Section \ref{ss:annotations}). After the checker annotated the frames, a second majority decision rule was applied over each five annotations per frame. 
%Majority was settled following the same steps as the first majority decision rule applied. 
The resolved labels between annotators increased to 92.6\% of the frames. A median filter was applied to decide the label for the remaining frames.

\begin{table*}[ht]
\caption{Agreement between annotators and checker for the annotated dataset}
\label{tab:agree}
\resizebox{\textwidth}{!}{%
\begin{tabular}{|c|c|c|c|c|c|c|c|c|c|c|c|c|c|c|c|c|c|c|c|c|c|c|}
\hline
Set \# & 1 & 2 & 3 & 4 & 5 & 6 & 7 & 8 & 9 & 10 & 11 & 12 & 13 & 14 & 15 & 16 & 17 & 18 & 19 & 20 & Mean & Std \\ \hline
\# Frames & 5197 & 5562 & 6317 & 6706 & 5340 & 6222 & 4861 & 6907 & 6433 & 6757 & 5295 & 6820 & 7312 & 8122 & 7396 & 6014 & 7130 & 8976 & 8689 & 4833 & - & -\\ \hline
%\begin{tabular}[c]{@{}c@{}}\% Agreement between \\ annotators\end{tabular} & 94.15 & 83.75 & 78.12 & 68.92 & 82.02 & 65.67 & 81.98 & 80.32 & 92.04 & 91.86 & 64.49 & 76.73 & 83.52 & 74.17 & 82.18 & 76.02 & 86.94 & 65.66 & 88.20 & 80.96 \\ \hline
\begin{tabular}[c]{@{}c@{}}\% Majority settled between \\ annotators\end{tabular} & 94.15 & 83.75 & 78.12 & 68.92 & 82.02 & 65.67 & 81.98 & 80.32 & 92.04 & 91.86 & 64.49 & 76.73 & 83.52 & 74.17 & 82.18 & 76.02 & 86.94 & 65.66 & 88.20 & 80.96 & 79.89 & 8.8245\\ \hline
%\begin{tabular}[c]{@{}c@{}}\% Agreement between \\ annotators and checker\end{tabular} & 98.19 & 94.70 & 90.87 & 85.71 & 94.14 & 85.26 & 95.23 & 94.90 & 98.91 & 97.60 & 82.85 & 90.62 & 93.34 & 93.70 & 92.39 & 91.57 & 96.06 & 87.32 & 96.05 & 92.72 \\ \hline
\begin{tabular}[c]{@{}c@{}}\% Majority settled\\ with checker\end{tabular} & 98.19 & 94.70 & 90.87 & 85.71 & 94.14 & 85.26 & 95.23 & 94.90 & 98.91 & 97.60 & 82.85 & 90.62 & 93.34 & 93.70 & 92.39 & 91.57 & 96.06 & 87.32 & 96.05 & 92.72 & 92.61 & 4.2085\\ \hline
\end{tabular}%
}
\end{table*}

\begin{figure*}[ht]
\centering%
\subfloat[Low\label{fig:low}]%
	{\includegraphics[width=0.3\linewidth]{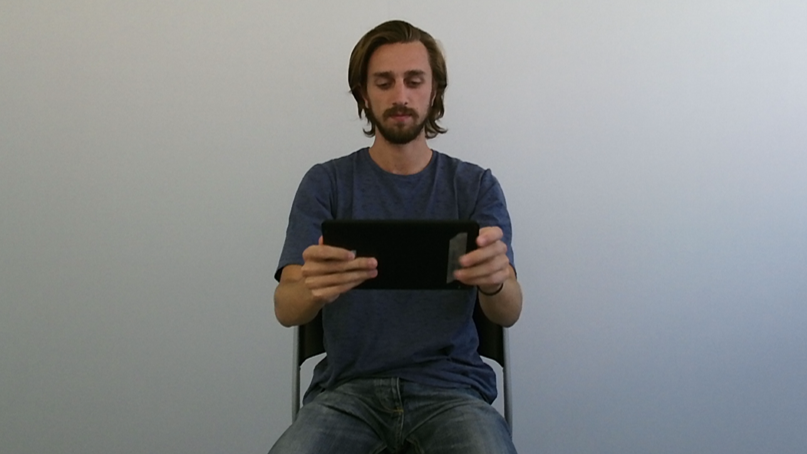}}%
    \hfill%
\subfloat[Mid\label{fig:mid}]%
	{\includegraphics[width=0.3\linewidth]{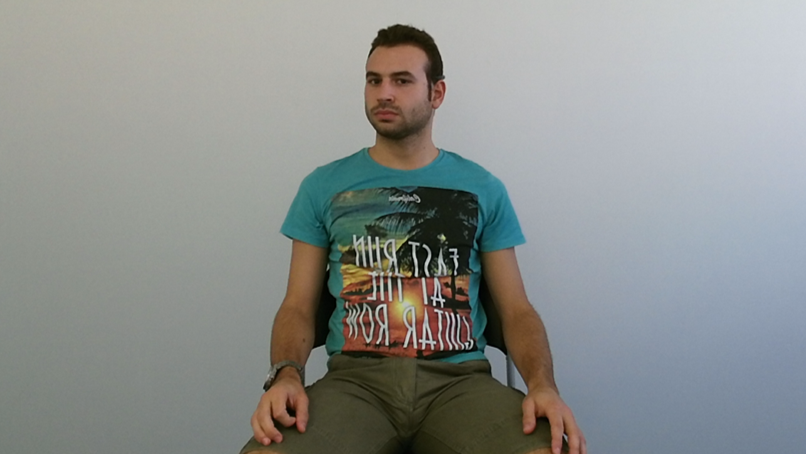}}%
    \hfill%
\subfloat[High\label{fig:high}]%
	{\includegraphics[width=0.3\linewidth]{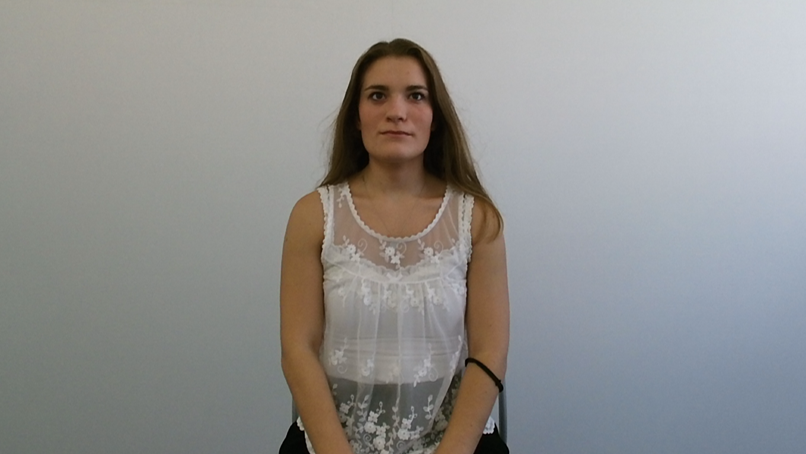}}%
\caption{Example of frames with the corresponding three label of attention level.}%
\label{fig:levels}
\end{figure*}
The distribution of the annotations in the dataset is as follows: the labels corresponding to low attention level occupy most of the labeled data with 70.8\% of annotations; mid and high attention level annotations are more or less equally distributed with a total of 15.5\% and 13.7\% of annotations over the total labeled dataset, respectively. Table \ref{tab:agree} shows the number of frames and the agreement between annotators before and after integrating the checker, for each subject or set of images. Fig. \ref{fig:levels} shows an example of three different frames extracted from the Pandora dataset with their corresponding final label of level of attention, after the agreement between annotators: Fig. \ref{fig:low} shows a subject manipulating a second screen, representing subjective low level of attention; Fig. \ref{fig:mid} shows a subject which head position is not aligned with the direction of the gaze, representing the subjective medium attention level (mid); and Fig. \ref{fig:high} shows a subject looking directly to the sensor, which clearly represents the subjective high attention level.

\section{\uppercase{Attention Level Estimation Methods}}\label{s:baseline}

% \todo[inline]{General: More consistency in the use of tenses. In my opinion you should use present tense (in most sections).}
% \todo[inline]{I think this is too detailed at this stage.}
% \begin{enumerate}
% \item Three classic machine learning algorithms: Support Vector Machines (\gls{svm}), Logistic regression (\gls{logit}), and Multilayer Perceptron (\gls{mlp}).
% \item Five different \gls{dnns} models.
% \end{enumerate}

In this section, we first introduce the proposed features and methods for attention level estimation. First, we explain the procedure for extracting the keypoints, we define the geometric features and present our approach to integrate depth. And second, we explain three simple machine learning methods and five  complex deep learning models.
% \todo[inline]{Revisit this paragraph: Feature selection (removed??), training/validation/test sets (moved to next section??), two different baselines (more??), performance (next section??). Probably needs a full re-writing.}

%We extract three different types of features: keypoints, geometric features, and depth-based.

% In this section, the different parts of the implemented algorithm are described. The steps include feature extraction, data processing, feature selection, and model learning and prediction. Figure represents the diagram of the implemented system for measuring and classifying attention level of a user.

% \begin{figure*}[!h]
%   %\vspace{-0.2cm}
%   \centering
%    \includegraphics[width=\textwidth]{Implementation}
%   \caption{Implementation diagram. \textbf{This figure needs a larger font size. Try to reduce whitespace.}}
%   \label{fig:diagram}

% \end{figure*}

% \begin{table}[b]
% \caption{Number and parts of the extracted body, face, and eye keypoints.}\label{tab:num_keypoints}
% \resizebox{\linewidth}{!}{%
% \begin{tabular}{|l|c|c|c|c|c|c|}
% \hline
% Keypoint & Nose & Neck & Shoulders & Eye centers & Eye contours \\ \hline
% Number & 1 & 1 & 2 (L/R) & 2 (L/R) & $6 \times 2$ (L/R) \\ \hline
% Part & Face & Body & Body & Face & Eye \\ \hline
% \end{tabular}%
% }
% \end{table}

\subsection{Feature Extraction}\label{ss:feature_extraction}

All keypoint coordinates are extracted using the publicly available OpenPose API \cite{hidalgo_openpose:_2018} from the RGB images of the annotated dataset. We use RGB color space since the original OpenPose model was trained using the same color space. For addressing the problem of attention estimation, we extract keypoints representing body and face parts of the user. Neck and right/left shoulder keypoints are extracted for describing the upper-body of the subject. Nose and eye center keypoints for the right and left eye are extracted for describing the face of the subject, and six eye contour keypoints for describing the right/left eyes. In cases when the API outputs multiple people for only one person (e.g. either a false positive detection or breaking the keypoints of one person into two), we find the similarity between two keypoint vectors and merge them if the missing keypoints are correlated. Since some of the keypoints from various frames can still be missing after this operation, we interpolate between the last detection and the most recent detection, for each keypoint, to create a continuous stream of detections.
% \todo{Last part violates assumptions of online/real-time computations, since we don't have future frames when testing. Make sure that either (1) we don't claim to present an online system, or (2) don't violate this, e.g. by simply predicting from previous frames instead of interpolating.}

% Applied to our data, this real-time multi-person system detects body and face part locations
%as two coordinates (according to the original source resolution); 
% and a confidence score of each detection normalized in the range [0,1]. 

% Figure \ref{fig:keypoints} shows an example of the representation of the keypoints in green. The image shows the keypoints for the neck, right and left eye, and right and left shoulder of a user.

% \begin{table}[h]
% \caption{This caption has one line so it is
% centered.}\label{tab:keypoints} \centering
% \begin{tabular}{|c|c|c|c|c|c|c|}
%   \hline
%   Part & Nose & Neck & RShoulder & LShoulder & REye & LEye\\
%   \hline
%   N$^{\circ}$ Keypoints & 1 & 1 & 1 & 6 contour & 6 contour \\
%   \hline
% \end{tabular}
% \end{table}

% \begin{figure}[h!]
% \centering
% \includegraphics[width=0.75\linewidth]{figures/keypoints}
% \caption{Graphical representation of keypoints. Original image extracted from Pandora dataset \cite{borghi_poseidon:_2017}.}
% \end{figure}

Before constructing our geometric features described below, we recalculate the position of each keypoint on each frame according to the coordinate system where the nose keypoint represents the origin. This way we can transform our features from being relative to the image coordinate system and instead construct them localized.

% \todo[inline]{I suggest to merge Figure 1 and 2, and show all distances and angles (if it doesn't get too messy). You could use color coding to show the different distances in $D$ (e.g. blue for $D_F$). Also, place the figure in the top or bottom of the page (don't use h).}

% \begin{figure}[!h]
% \centering
% \includegraphics[width=\linewidth]{figures/keypoints}
% \caption{Neck, eyes, and shoulders keypoints represented in green.}
% \label{fig:keypoints}
% \end{figure}

%For false positive detections, we filter the results to use the detection that has more available keypoints. False positive detections have usually a very few keypoints available. We re-define the keypoints into a feature space as follows: nose, neck, right eye, left eye, right shoulder, left shoulder, refined right eye (pupil and six contour keypoints), and refined left eye (pupil and six contour keypoints). We make the interpolation for zero keypoints. The coordinates of the keypoints are recalculated according to the nose keypoint, which is settle as origin. In this way, all keypoints will be global coordinate independent.
% We filter the resulting keypoint sets for false positive detections, and transform the keypoints such that the nose is located in the origin. In this way, keypoints are not affected by different locations of users in the images. We solve missing keypoints by interpolating positions between frames. 

The geometric features were inspired by Zhang \textit{et al.} \cite{zhang_fusing_2018} and are constructed based on the extracted keypoints. They consist of a set of relations between keypoints. Two types of geometric features were defined:
\begin{enumerate}
\item A set, $D$, of Euclidean distances between keypoints.
\item A set, $A$, of angles between two unit vectors, each of which represents a line linking two consecutive keypoints.
\end{enumerate}
The first set of geometric features is described by four types of distances: 
\begin{equation}
D = \{ D_F, D_{BF}, D_{LE}, D_{RE} \},
\end{equation}
where $D_F$ corresponds to the Euclidean distances between face keypoints and is defined by:
\begin{equation}
D_F = \{||\pmb f_j - \pmb f_i||\ | \ i, j \in F,\ i<j \},
\end{equation}
with $F$ being the face keypoints, and $\pmb f_i$ the coordinate of the $i^{th}$ face keypoint.
$D_{BF}$ corresponds to the Euclidean distances between face and body keypoints and is defined by:
\begin{equation}
D_{BF} = \{||\pmb f_j - \pmb b_i||\ | \ i \in B,\ j \in F\},
\end{equation}
where $B$ are the body keypoints and $\pmb b_i$ the coordinate of the $i^{th}$ body keypoint.
$D_{LE}$ corresponds to the Euclidean distances between left eye center keypoint to each six contour keypoints of the left eye and they are defined by:
\begin{equation}
D_{LE} = \{||\pmb e_i - \pmb c_{L}||\ | \ i \in LE \},
\end{equation}
with $LE$ being the contour keypoints of the left eye, $\pmb e_i$ the coordinate of the $i^{th}$ contour keypoint, and $\pmb c_{L}$ the coordinate of the left eye center.
Similarly, $D_{RE}$ corresponds to the Euclidean distances between right eye center keypoint to each six contour keypoints of the right eye.

For the definition of the set, $A$, we compute three unit vectors. One between nose and neck:
\begin{equation}
\pmb v_1 = \frac{\pmb f_{nose} - \pmb b_{neck}}{||\pmb f_{nose} - \pmb b_{neck}||}.
\end{equation}
And two between neck and left/right shoulders, respectively:
\begin{equation}\small
\pmb v_2 = \frac{\pmb b_{Lshoulder} - \pmb b_{neck}}{||\pmb b_{Lshoulder} - \pmb b_{neck}||},
\pmb v_3 = \frac{\pmb b_{Rshoulder} - \pmb b_{neck}}{||\pmb b_{Rshoulder} - \pmb b_{neck}||}.
\end{equation}
We then describe the set as:
\begin{equation}
A = \left\{ \arccos \left( \pmb v_1^\text{T} \pmb v_2 \right),\ \arccos \left( \pmb v_1^\text{T} \pmb v_3 \right) \right\},
\end{equation}
which are two angles between neck to nose vector and vectors from neck to left/right shoulders, respectively.
% Given a set of three keypoints

% \begin{equation*}
% \pmb K = \{ \pmb K_{1}, \pmb K_{2}, \pmb K_{3}  \}
% \end{equation*}

% corresponding to the neck, nose, and right/left shoulder, respectively. We define the orientation from neck to nose and from neck to right/left shoulder as

% \begin{equation*}
% \pmb O_{2}(\pmb K_{1}, \pmb K_{2}) = unit(\overrightarrow{\pmb K_{1}, \pmb K_{2}})
% \end{equation*}

% \begin{equation*}
% \pmb O_{2}(\pmb K_{1}, \pmb K_{3}) = unit(\overrightarrow{\pmb K_{1}, \pmb K_{3}})
% \end{equation*}

% Then we define the angles between the unit vector neck-nose and the unit vector neck-right/left shoulder as

% \begin{equation*}
% \theta = \arccos( \pmb (O_{1})^{T} \odot \pmb O_{2}),
% \end{equation*}

% where $\odot$ denotes the dot product.

\begin{figure}
\centering
\includegraphics[width=\linewidth]{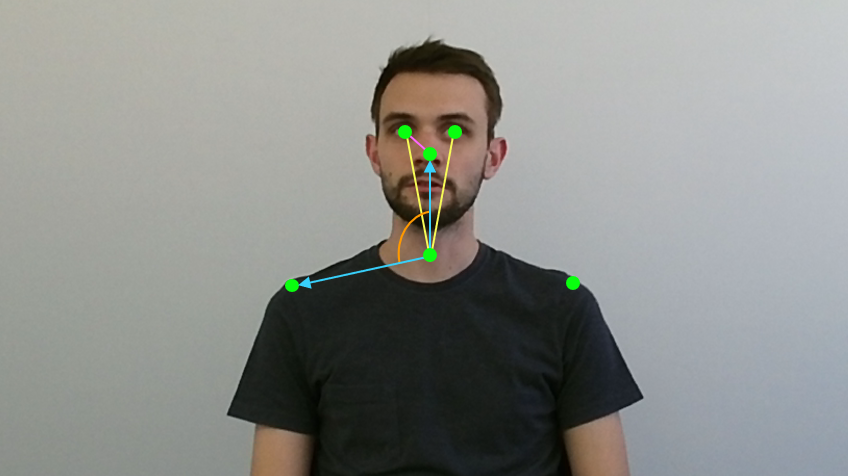}
\caption{Graphical representation of keypoints and geometric features in colors: green for keypoints, pink for $D_{F}$, yellow for $D_{BF}$, blue for vectors used to compute the angles (orange). Original image extracted from Pandora dataset \cite{borghi_poseidon:_2017}.}
\end{figure}

Research within computer vision has shown that, in many applications, when depth information is included along with RGB imagery, the extra modality can improve the performance of the system \cite{li_real_2009,molina_real-time_2015}. We extract depth information on a point of interest basis. This process covers finding positions on the original RGB image, where depth information is to be extracted using the already computed keypoints as keys.
%\todo[inline]{I think this is too trivial.}
%We extract information for the average depth at each already extracted keypoints and for its $n$ neighboring pixels. The depth at each point is calculated as:
%
%\begin{equation*}
%D_{ij} = \frac{1}{n} \sum_{k=1}^{n}D_{ijk},
%\end{equation*}
%
%where $n=9$, and $D_{ij}$ is the average depth (pixel intensity) pf the $j$'th extracted keypoints, for the $i$'th sample in the data.
By taking the average depth in a 3x3 area around each position, we alleviate the problem of possible small positional errors in the extracted keypoints. Note that the extraction of depth pixel intensity information is done after interpolating the missing keypoint coordinates, but before recalculating the keypoint coordinates according to the Nose keypoint. Once all features are extracted, we standardize them.

\subsection{Attention Level Estimation}

In this section, we present several methods for attention estimation with subjective attention annotations, which can serve as baselines for further development. We use our own baseline for attention level estimation, since we find that there are no adequate baselines at the present and we consider our novel attention level estimation framework and subjective labeling process a pioneer work.

% \begin{enumerate}
% \item Three classic machine learning algorithms: Support Vector Machines (\gls{svm}), Logistic regression (\gls{logit}), and Multilayer Perceptron (\gls{mlp}).
% \item Five different Deep Neural Networks (\gls{dnns}) models.
% \end{enumerate}

% \todo[inline]{Use this section for describing the methods (and save cross-validation for experiments, see my comment further down). We need more details on the implementation. What parameters have you used? Is MLP different from DNN??}

\begin{figure*}[t]
\centering%
\subfloat[Early\label{fig:early}]%
	{\includegraphics[width=0.3\linewidth]{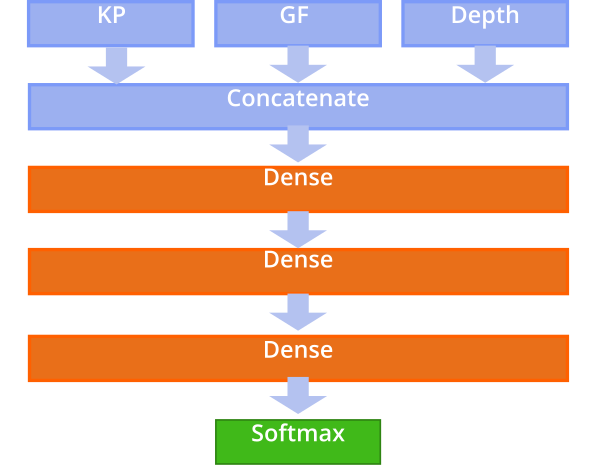}}%
    \hfill%
\subfloat[Fully-connected\label{fig:fc}]%
	{\includegraphics[width=0.3\linewidth]{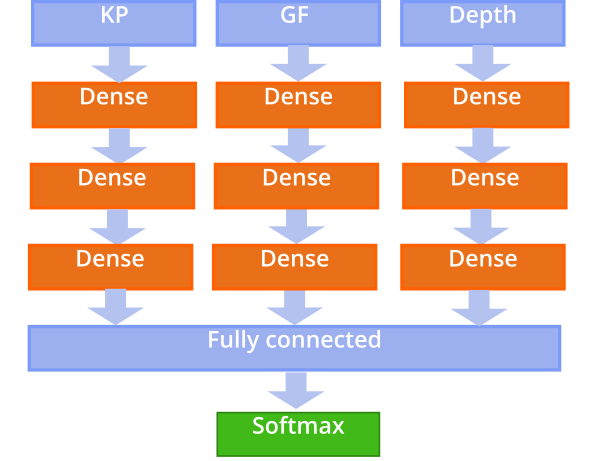}}%
    \hfill%
\subfloat[Late\label{fig:late}]%
	{\includegraphics[width=0.3\linewidth]{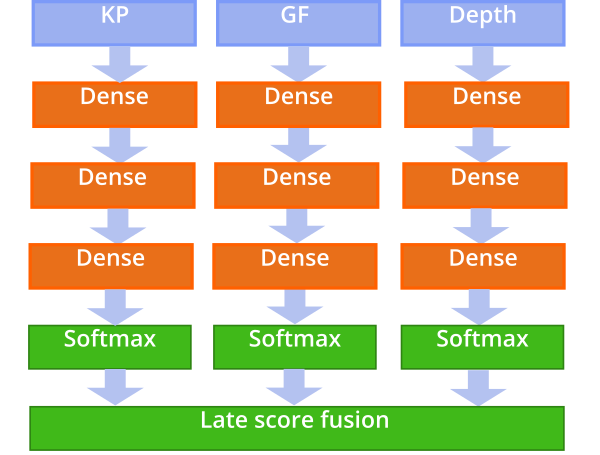}}%
\caption{DNN models and methods for fusing keypoints (KP), geometric features (GF) and depth.}%
\end{figure*}

Before training \gls{dnn} models for attention level classification, we train and evaluate three classic machine learning models, to set up a baseline for our more complex models’ performance. We train Support Vector Machine (SVM), Logistic Regression (Logit) and a single hidden layer Multilayer Percepton (MLP) for multiclass classification.

% \todo[inline]{It's not necessary to show the results of each fold. Instead just show the average and the standard deviation (e.g. in parentheses after the average). Example for SVM: 0.6121 (0.0037).}

\subsubsection{Deep Neural Network (\gls{dnn}) Models}\label{ss:dnn_models}
% \todo[inline]{If we don't include the feature selection this paragraph is not relevant.}
% The results from feature selection might not always yield the best possible combinations of features since the correlation between the features themselves might differ. As an example, consider a possible subset of selected features $f_{ijk}$, where $f_{i}$, $f_{j}$, and $f_{k}$ represent the selected geometric features of keypoints, distances, and angles, respectively. Then $f_{ijk}$ is the concatenation of all selected features in a feature vector. In this paper, three different fusion approaches are implemented.

Using all of our feature space as a whole, as the input to our \gls{dnn} models (``early" fusion, see Fig.~\ref{fig:early}), might not always yield the best possible result. Other studies made their efforts on fusing multiple keypoints and geometric features \cite{zhang_fusing_2018}, and argue that fusing these features together at a later stage of the network - either by training a fully connected layer on top of different dense layer streams (``fully-connected" fusion, see Fig.~\ref{fig:fc}) or combining the softmax decisions (``late" fusion, see Fig.~\ref{fig:late}) - can further increase the overall classification accuracy. To further evaluate our extracted features, we decided to follow the three feature fusion frameworks, explained below. Note that in all models where we apply fully-connected or late fusion, we handle keypoints, geometric features and/or depth in streams of separate dense layers. Further on, in this paper, these streams are noted as \textit{feature streams}.

% \todo[inline]{The descriptions are a bit blurry (we need more specifics, e.g. \# of layers, etc.), and I don't think a new reader would fully understand the different approaches. I would suggest making a simple figure showing the three different methods (in vector graphics ;-) ).}
\textbf{Early fusion:} the keypoints and geometric features (when having two modalities) and depth (three modalities) are concatenated and given as input to the \gls{dnn} models during training and inference.

% \begin{figure}[h!]
% \centering
% \includegraphics[width=0.7\linewidth]{figures/early.png}
% \caption{"Early" fusion method}\label{fig:early}
% \end{figure}

\textbf{Fully-connected feature fusion:} the degree of how much each feature subset contributes to the classification task at hand is learned through different information streams based on the input. This is possible to be learned through trainable weights of a \gls{dnn}. Each stream represents layers of densely connected neurons that are kept separately for each stream and are only connected to a fully connected layer at a later stage, before classification. This way the different layers of each stream can learn unbiased information by the other stream, but share information through the neurons of the fully connected layer, before classification.

% \begin{figure}[h!]
% \centering
% \includegraphics[width=0.7\linewidth]{figures/fc.png}
% \caption{"Fully-connected" fusion method}\label{fig:fc}
% \end{figure}

% \todo[inline,color=blue]{Perhaps it would be a good idea to formulate how the fusion is achieved.}

\textbf{Late fusion:} this approach integrates common meaning representations derived from different modalities into a combined final interpretation, by utilizing separate classifiers that can be trained independently. The final decision is reached by combining the partial outputs of the unimodal classifiers, by either taking the maximum (maximum fusion) or the average scores of all feature stream specific softmax layers (average fusion) train a set of learnable weights on top of each softmax layer and attaching a new softmax layer on top of these weights, to reach the final classification scores \cite{kuang_fruit_2016}.

Each of the dense layers have 256 and fully connected layers have 64 hidden units, with ReLU activations. The learning rate was initially set to 0.0001 for the Adam optimizer. In case of fully-connected fusion and weighted score fusion, we train the different feature streams separately. When we train the fully connected fusion models, the softmax layers of each stream are detached and connected to a fully connected layer which is further connected to its own softmax layer. We freeze the weights of each feature stream and only update the weights of the new layers during backpropagation. With this procedure, we can deny the bias in each stream caused by the parallel ones. We train our late fusion models in the same fashion, except we keep the softmax layers, and attach a new custom layer that operated on the decision of each stream.

\section{\uppercase{Experimental Results and Discussion}} \label{sec:experiments}

% \todo[inline]{I think we should make an experiments section here. This section would describe how we tested, e.g. using cross-validation, and show the results (try to combine Table 5 and 6?). That would give a better flow where we can focus on the methods in section 4 (4.1: Feature extraction, 4.2: Attention estimation, 4.2.1: Classic [...], 4.2.2: DNN). Make sure to report the standard deviation (or variance) of the results together with the average of the cross-validation results.}

Four-fold cross-validation were used for evaluating the presented algorithms. The final accuracy is obtained as the average of all the accuracies of all four models from each algorithm. Table \ref{tab:results} shows the performance of the classic machine learning algorithms over the dataset alongside the performance of the different \gls{dnn} models for each modality in terms of accuracy.

% \begin{table}[h!]
% \caption{Performance of the classic machine learning algorithms}
% \label{tab:classic} 
% \centering
% \resizebox{0.75\linewidth}{!}{%
% \begin{tabular}{|c|c|c|}
% \hline
% Model & Accuracy & Standard deviation \\ \hline
% SVM & 0.6121 & 0.0043 \\ \hline
% Logit & 0.5887 & 0.0030 \\ \hline
% MLP & 0.6314 & 0.0155 \\ \hline
% \end{tabular}%
% }
% \end{table}

% \begin{table}[h!]
% \caption{Performance of the different \gls{dnn} models}
% \label{tab:dnnper}
% \resizebox{\linewidth}{!}{%
% \begin{tabular}{|c|c|c|c|}
% \hline
% \multicolumn{2}{|c|}{Fusion model} & Two modalities & Three modalities \\ \hline
% \multicolumn{2}{|c|}{Early} & 0.7293 & 0.7547 \\ \hline
% \multicolumn{2}{|c|}{Fully-connected} & 0.7781 & 0.8002 \\ \hline
% \multirow{3}{*}{Late} & Average & 0.7196 & 0.7169 \\ \cline{2-4} 
%  & Maximum & 0.7161 & 0.7075 \\ \cline{2-4} 
%  & Weighted & 0.7236 & 0.7234 \\ \hline
% \end{tabular}%
% }
% \end{table}

% Please add the following required packages to your document preamble:
% \usepackage{multirow}
% \usepackage{graphicx}
% Please add the following required packages to your document preamble:
% \usepackage{multirow}
% \usepackage{graphicx}
\begin{table}[b]
\caption{Performance of the classic algorithms and DNN models with two (KP and GF) and three (KP, GF and depth) modalities.}
\label{tab:results}
\resizebox{\linewidth}{!}{%
\begin{tabular}{|c|c|c|c|c|}
\hline
\multicolumn{3}{|c|}{Model} & Accuracy & Standard deviation \\ \hline
\multicolumn{3}{|c|}{SVM} & 0.6121 & 0.0043 \\ \hline
\multicolumn{3}{|c|}{Logit} & 0.5887 & 0.0030 \\ \hline
\multicolumn{3}{|c|}{MLP} & 0.6314 & 0.0155 \\ \hline
\multicolumn{2}{|c|}{\multirow{2}{*}{Early fusion}} & KP + GF & 0.7293 & \multirow{10}{*}{(0.0020-0.0040)} \\ \cline{3-4}
\multicolumn{2}{|c|}{} & KP + GF + Depth & 0.7547 &  \\ \cline{1-4}
\multicolumn{2}{|c|}{\multirow{2}{*}{Fully-connected fusion}} & KP + GF & 0.7781 &  \\ \cline{3-4}
\multicolumn{2}{|c|}{} & KP + GF + Depth & 0.8002 &  \\ \cline{1-4}
\multirow{6}{*}{Late fusion} & \multirow{2}{*}{Average} & KP + GP & 0.7196 &  \\ \cline{3-4}
 &  & KP + GF + Depth & 0.7169 &  \\ \cline{2-4}
 & \multirow{2}{*}{Maximum} & KP + GF & 0.7161 &  \\ \cline{3-4}
 &  & KP + GF + Depth & 0.7075 &  \\ \cline{2-4}
 & \multirow{2}{*}{Weighted} & KP + GF & 0.7236 &  \\ \cline{3-4}
 &  & KP + GF + Depth & 0.7234 &  \\ \hline
\end{tabular}%
}
\end{table}

\begin{figure}[t]
\centering
\includegraphics[width=0.9\linewidth]{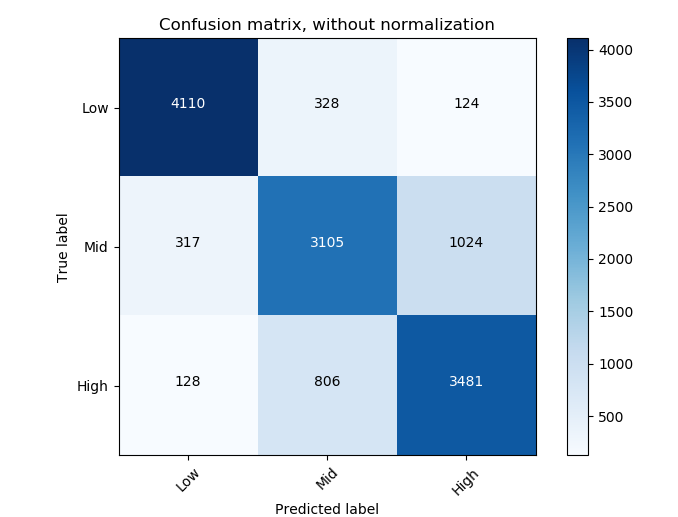}
\caption{Confusion matrix of the results of the fully-connected fusion of keypoints, geometric features and depth. Darker colors correspond to a larger number of classifications.}\label{fig:fc_conf}
\end{figure}

% \todo[inline]{Why not show the results in a single table? Put standard deviation of DNN results as well. It should also be possible to tell from the table what "two modalities" and "three modalities" refer to, i.e. what are those modalities?}
From our experiments, we find that including depth information among the input features is mostly beneficial. Table \ref{tab:results} shows that both the early and fully-connected fusion of keypoints, geometric features and depth outperforms the fusion of keypoints and geometric features. The three different versions of late fusion models did only make a marginal difference between the same results. From the results of our best performing model (fully-connected fusion with three modalities, see Table \ref{tab:results}) we find that most false negative and false positive classifications are performed regarding ``mid" attention level. Fig. \ref{fig:fc_conf} shows that more than 60\% of the false classifications happen along the ``mid" label. This result can be best explained by the nature of the annotations. Although in most cases ``low" and ``high" attention level was clearly decided by the majority votes of the annotators, ``mid" attention level was mostly decided by the final agreement rule or the checker's annotations. This meant that most images which caused confusion or a difficult decision for the annotators fell under the ``mid" attention label, resulting in the most difficult and versatile class of all three. It is also important to note that when one of the DNN models outperformed a previous one, the performance change was best visible on the misclassification rate of the ``mid" class.

The results in Table \ref{tab:per_stream} show how accurately individual streams estimate single attention level classes (accuracy per individual class label). It is clearly visible that keypoints perform the estimation of low and mid attention level the most accurate, however geometric features are the best at estimating high attention level. Although depth is a overall low accuracy descriptor of attention, the results from Table \ref{tab:results} justify the inclusion of depth information. These results show fusion (on any level) of different modalities can help to increase the model's overall performance for the task at hand. Table \ref{tab:results} shows that late fusion's performance was always inferior compared to the other two versions of fusion. This is best explained by late fusion only being a powerful tool when it is introduced to modalities that are represented differently in the input data. This difference can refer to difference in temporal aspects or representation (e.g images and numeric data or sound). Neither of the feature subsets and modalities that were fused together differ on a temporal aspect and the way these features or modalities are represented in the feature space, which explains the inferior performance of all late fusion models. To validate these results, we introduced a new fusion model, where the geometric features were fused together in a fully-connected model, and fused together with the depth features in a weighted late fusion model. The results of the new fusion can not over perform the results of the best fully connected fusion model either. The better performance of the fully connected fusion model can be best explained by the nature of the method. Since all streams are kept separated for the early stages of the network, the learned weights inside the stream specific layers are less correlated to each of the fusion streams. However, the classification error is propagated back to each of these layers, during global optimization. Therefore, the model can still adjust its learned parameters according to the other separated streams. This way, the learned information is only limited throughout the early stages of training.
\begin{table}[t]
\caption{Results of how accurately individual streams estimate single attention level classes (accuracy per individual class label).}
\label{tab:per_stream}
\resizebox{\linewidth}{!}{%
\begin{tabular}{c|c|c|c|}
\cline{2-4}
\multicolumn{1}{l|}{} & \multicolumn{3}{c|}{Accuracy per individual class label} \\ \cline{2-4} 
 & Low & Mid & High \\ \hline
\multicolumn{1}{|c|}{Keypoints} & 0.8938 & 0.7470 & 0.7807 \\ \hline
\multicolumn{1}{|c|}{Geometric Features} & 0.8909 & 0.7434 & 0.7886 \\ \hline
\multicolumn{1}{|c|}{Depth} & 0.7764 & 0.6704 & 0.6745 \\ \hline
\end{tabular}%
}
\end{table}

\section{\uppercase{Conclusion}}\label{s:conclusion}

Estimating attention level of a user is a very challenging task. The majority of methods rely on describing attention by measuring the \gls{vfoa} of a user with combination of head pose estimation. The development of datasets in the field have been at the focus point in research over the past years. However, existing datasets either do not include annotations or rely on objective annotations depending on the direction of the eye gaze and head pose of a user. In this paper, we proposed a novel approach towards estimating attention level of a user with subjective annotation levels by evaluating geometric features. We hand-labeled over 100,000 images of the Pandora dataset with three levels of subjective annotations, using five participants. The objective of the labeling process was to label attention level of data based solely on personal feelings and opinion, which we believe is beneficial for tasks such as estimating the attention level of a user, to incorporate the subjective nature of attention itself. We further set up baseline results of attention level estimation, using our annotations and different deep learning fusion models. Our best achieved accuracy was 80.02\% on attention level estimation. 
As a future work, we consider labeling the dataset for other applications in the field related to attention, such as the \gls{vfoa} (looking at the TV or not) of each person, frame-by-frame. Shift of attention labels can also be added, e.g annotating the frames where the attention shifts from low to mid, mid to high and high to low.

% \section{\uppercase{Future Work}}\label{s:future_work}

% Future development of the system should be focused on optimizing the dataset. This could potentially be achieved by creating a specific dataset for addressing the problem of classifying attention level of a user in the living room with subjective attention labels. To this end, we collected our own dataset: Granade. The Granade dataset captured the body of two different subjects watching television in a living room with a Microsoft Kinect for Xbox 360. It contains 33.060 RGB and depth (1280x720 pixels) images. Future work includes hand-labeling the Granade dataset following the same process as explained in \ref{ss:annotations}.

\bibliographystyle{apalike}
% \bibliography{17gr949}}
\end{document}